\documentclass[letterpaper, 10 pt, conference]{ieeeconf}  

\IEEEoverridecommandlockouts                              

\usepackage{graphics} 
\usepackage{epsfig} 
\usepackage{mathptmx} 
\usepackage{times} 
\usepackage{amsmath} 
\usepackage{amssymb}  
\usepackage{graphicx}
\usepackage{mathtools}
\usepackage{algorithmic}
\usepackage{cite}
\usepackage{xcolor}

\title{\LARGE \bf
Resilient Navigation for Autonomous Farm Robots by Leveraging Jerk-Augmented Models with IMU-Only Disturbance Rejection}

\author{Batu Candan$^{1}$, Mohammed Atallah$^{2}$, Simone Servadio$^{3}$ and Saeed Arabi$^{4}$
\thanks{*The work of this project has been funded by the Seed Grant Award for Digital and Precision Agriculture at Iowa State University under grant number PG114523.}
\thanks{$^{1}$Batu Candan, PhD Candidate,
         Iowa State University, Ames, IA 50011, USA,
        }%
\thanks{$^{2}$Mohammed Atallah, PhD Candidate,
         Iowa State University, Ames, IA 50011, USA,
        }%
\thanks{$^{3}$Dr. Simone Servadio, Assistant Professor,
         Iowa State University, Ames, IA 50011, USA,
        }%
\thanks{$^{4}$Saeed Arabi is the founder and software lead, Salin247,
        Boone, IA 50036, USA
        }%
}

\begin{document}

\maketitle
\thispagestyle{empty}
\pagestyle{empty}

\begin{abstract}

Precise state estimation for navigation of autonomous agricultural robots is often compromised by sensor outages (GNSS/LiDAR/Visual) and high-frequency vibrations inherent in off-road environments. This paper proposes a robust navigation algorithm based on a jerk-augmented Extended Kalman Filter (EKF) integrated with a Multiple Tuning Factor (MTF) adaptation method. Unlike standard EKF approaches that assume constant measurement noise, our method dynamically adjusts the measurement covariance matrix in real-time, allowing the system to cope with sudden disturbances and sensor outliers. We evaluate the algorithm using real-world data from a Salin247 autonomous robot. Results demonstrate that jerk-augmentation combined with MTF adaptation significantly reduces 3D position Root Mean Square Error (RMSE) compared to baseline EKF models, providing superior dead-reckoning capabilities.

\end{abstract}

\section{INTRODUCTION}

The growing global demand for sustainable agricultural production, coupled with labor shortages and the need for high operational precision, has accelerated the adoption of autonomous agricultural vehicles and robots. These systems are increasingly tasked with operations such as weeding, spraying, harvesting, and monitoring, where precise navigation is essential. The fundamental challenge, however, remains the reliable and accurate navigation of such vehicles in complex, dynamic, and unstructured agricultural environments \cite{b1}. To overcome these constraints, researchers have explored several autonomous navigation strategies, including GNSS, LiDAR, inertial sensors, and camera-based machine vision \cite{b2}. However, those systems are susceptible to signal degradation or loss due to tree canopies, farm structures, or adverse atmospheric conditions. Moreover, agricultural terrain is inherently uneven. The resulting mechanical vibrations and impulsive shocks create noisy inertial data, especially on accelerometers, which can mislead standard Kalman filtering applications \cite{b3}. 

Many studies focus on integrating low-cost GNSS/IMU systems using standard 6-state or 9-state EKF models \cite{Farahan, Kaczmarek}. While these provide a baseline for mild conditions, they rely heavily on the assumption of constant velocity or constant acceleration between IMU samples. As demonstrated in our results, these models suffer from lag and cumulative error when the robot encounters impulsive terrain disturbances or sudden torque changes, as they lack the higher-order derivative, jerk, necessary to model the rate of change of acceleration. Moreover, to handle the unpredictable nature of farm environments, several researchers have proposed Adaptive EKFs (AEKF). However, most existing adaptive methods use a scalar tuning factor that scales the entire measurement covariance matrix ($\textbf{R}$) or process noise matrix ($\textbf{Q}$) uniformly \cite{b4, b5, b12, b6}. Those recent adaptive methods use various different strategies. For instance, Liu et al. \cite{sagehusa} implemented innovation-based adaptive estimation methods \cite{innov1, innov2, innov3}, such as the Sage-Husa filter, to statistically tune the measurement noise covariance.  

Other researchers adopted federated Kalman Filter (FKF) methodology to improve fault tolerance. These systems utilize an information distribution factor, often denoted as $\beta_i$, to distribute the fused global information back \cite{rice}. However, again, the factor $\beta_i$ scales the entire covariance matrix of a local filter uniformly. Consequently, it lacks the ability required to handle axis-specific disturbances. Similarly, fuzzy-logic based adaptation schemes have been explored to dynamically adjust filter gain based on heuristic rules \cite{fuzzy1, fuzzy2, fuzzy3}. However, a critical limitation in these approaches is their reliance on a scalar tuning. These methods scale the entire measurement covariance matrix uniformly. These ``all-or-nothing" approaches are sub-optimal for multi-axis agricultural robots; for example, a vertical impact caused by a soil rut may corrupt the vertical accelerometer data, but the lateral data often remains valid. Scaling the entire matrix forces the filter to discard useful lateral information, reducing overall precision.

While some methodologies attempt to map noise effects onto the space domain or divide uncertainties across multiple models to improve trackability \cite{b13, b14}, they often lack directional specificity. To address these limitations, this work proposes a resilient navigation framework tailored for agricultural robotics. First, our approach incorporates jerk into the state vector to accurately model the rate of change of acceleration, mitigating the lag caused by sudden terrain disturbances. Second, we introduce an axis-specific,  measurement noise covariance adaptation based on the analysis of residuals at each time step. By dynamically and independently adjusting the covariance along each axis, the proposed filter isolates localized disturbances without discarding valid sensory data from the unaffected axes, thereby maintaining robust and precise navigation in unstructured environments.

\section{PRELIMINARIES}
This section outlines the foundational components of the proposed navigation framework. We first describe the autonomous agricultural platform used for validation, followed by the mathematical formulation of the attitude representation and sensor models. Finally, we derive the process and measurement models for the baseline KF structure, identifying the specific limitations in handling external accelerations that our proposed method addresses.

\subsection{Salin247 Autonomous Farming Robot}
\begin{figure}[htbp]
\centerline{\includegraphics[width=0.5\textwidth]{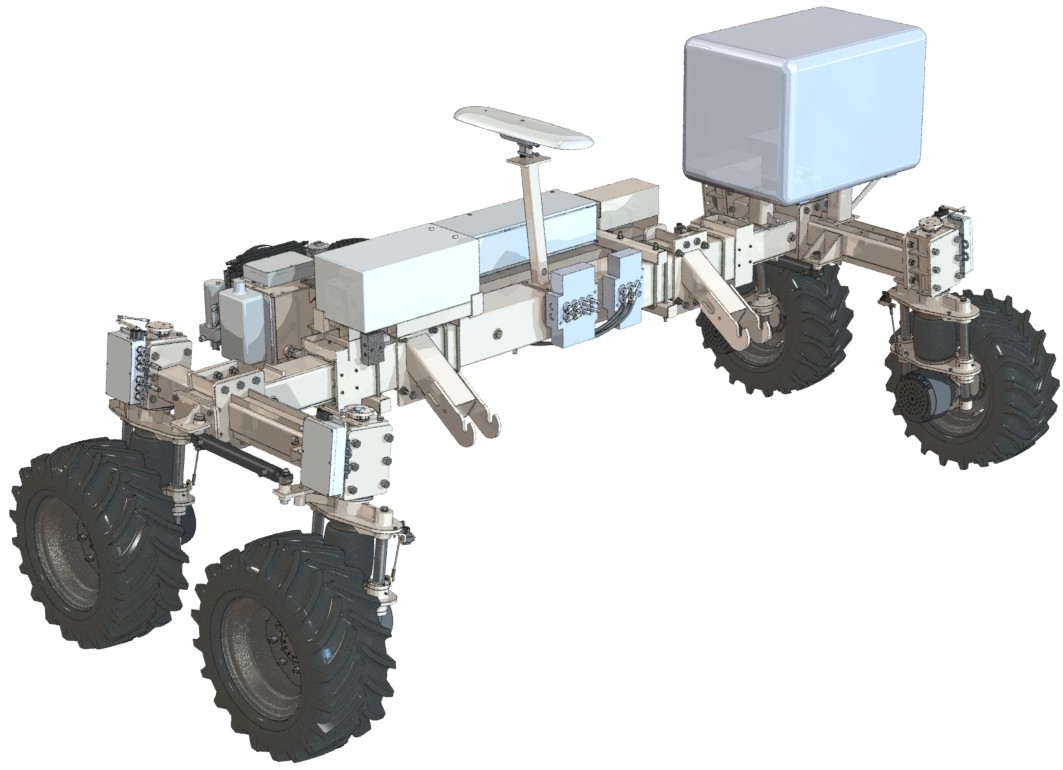}}
\caption{Salin247 autonomous agricultural robot \cite{b_salin}}
\label{fig11}
\end{figure}

The experimental validation was conducted using the Salin247 autonomous agricultural robot, a modular platform designed to address soil compaction and labor shortages in precision farming. As illustrated in Fig. \ref{fig11}, the vehicle consists of a light-weight, fully autonomous, electric-powered chassis that attaches easily to customizable implements such as planters, cultivators, and spot-sprayers. Unlike traditional heavy tractors, the Salin247 platform utilizes a distributed propulsion system with four independently driven electric track units. This 4-wheel independent drive (4WID) configuration allows for precise differential steering and high torque delivery at low speeds, which is essential for maintaining traction on uneven agricultural terrain \cite{b_salin}.

\subsection{Attitude Representation}
Reliable navigation for autonomous agricultural platforms requires an accurate representation of the vehicle's orientation. Common attitude representations include Euler angles and quaternions; however, in this study, we use the Direct Cosine Matrix (DCM) representation to avoid the singularities associated with Euler angles during steep maneuvers on uneven terrain. Furthermore, for 2-D dominant farm navigation, DCM is more parameter-efficient than quaternions, which possess redundant parameters. Defining $S$ and $I$ as the sensor and inertial coordinate frames, let ${}^{I}_{S}\boldsymbol{\Omega}$ (simplified here as $\boldsymbol{\Omega}$) denote the DCM that relates the sensor frame to the inertial frame. Adopting the standard Z-Y-X Euler angle rotation sequence, we first define the individual Direction Cosine Matrices (DCMs). The rotations around the X, Y, and Z axes are parameterized by the roll ($\gamma$), pitch ($\beta$), and yaw ($\alpha$) angles, respectively. Letting $c$ and $s$ abbreviate the cosine and sine functions, the fundamental rotation matrices are formulated as:

\begin{align}
\textbf{R}_z(\alpha) &= \left[\begin{matrix}c\alpha&-s\alpha&0\\s\alpha&c\alpha&0\\0&0&1\end{matrix}\right], \\[6pt]
\textbf{R}_y(\beta) &= \left[\begin{matrix}c\beta&0&s\beta\\0&1&0\\-s\beta&0&c\beta\end{matrix}\right], \\[6pt]
\textbf{R}_x(\gamma) &= \left[\begin{matrix}1&0&0\\0&c\gamma&-s\gamma\\0&s\gamma&c\gamma\end{matrix}\right].
\end{align}

The composite rotation matrix, $\boldsymbol{\Omega}$, is formed by multiplying these individual matrices in the $\textbf{R}_z \textbf{R}_y \textbf{R}_x$ sequence as follows. Note that while this formulation adopts the standard Z-Y-X (or 3-2-1) Euler angle rotation sequence, it is just one of twelve possible rotation orders, which can be selected based on the specific kinematic requirements of the system.

\begin{equation}
\boldsymbol{\Omega}=\textbf{R}_z(\alpha)\textbf{R}_y(\beta)\textbf{R}_x(\gamma)
\end{equation}

The two-axis attitude or tilt angles (roll and pitch) required for gravity compensation can be extracted solely from the third row of $\boldsymbol{\Omega}$.
\begin{equation}
\gamma=\tan^{-1}\left(\frac{\boldsymbol{\Omega}_{32}}{\boldsymbol{\Omega}_{33}}\right),
\end{equation}
\begin{equation}
\beta=\tan^{-1}\left(\frac{-\boldsymbol{\Omega}_{31}}{\sqrt{\boldsymbol{\Omega}_{32}^2+\boldsymbol{\Omega}_{33}^2}}\right). 
\end{equation}
Letting $\boldsymbol{\Omega}_{ij}$ denote the $(i,j)$-th element of $\boldsymbol{\Omega}$, we define the state vector $\mathbf{x}$ for this formulation using the matrix's third row:
\begin{equation}
\mathbf{x}=\boldsymbol{\Omega}^T\mathbf{e}=\left[\begin{matrix}\boldsymbol{\Omega}_{31}\\\boldsymbol{\Omega}_{32}\\\boldsymbol{\Omega}_{33}\\\end{matrix}\right],   
\end{equation}
where $\textbf{e}=\begin{bsmallmatrix} 0 & 0 & 1 \end{bsmallmatrix}^T$. This specific Euler sequence introduces a kinematic singularity when $\boldsymbol{\Omega}_{31}=\pm1$, which is avoided provided the platform's pitch angle remains bounded away from $\pm90^{\circ}$.

\subsection{Sensor Models}
The raw measurements from the gyroscope ($\mathbf{y}_G$) and accelerometer ($\mathbf{y}_A$) are given by:
\begin{equation}
\label{eq5}
\mathbf{y}_\mathbf{G}={_{}^S}\boldsymbol{\omega}+\boldsymbol{n}_G,
\end{equation}
\begin{equation}
\label{eq6}
\mathbf{y}_\mathbf{A}={_{}^S}\mathbf{a}+{_{}^S}\mathbf{g}+\boldsymbol{n}_A,
\end{equation}
where the $S$ superscript denotes the sensor frame. The terms ${_{}^S}\boldsymbol{\omega}$ and ${_{}^S}\mathbf{a}$ represent the true angular velocity and external acceleration, respectively, while the gravity vector is represented with ${_{}^S}\mathbf{g}$. The variables $\boldsymbol{n}_G$ and $\boldsymbol{n}_A$ denote zero-mean white Gaussian sensor noises, which are uncorrelated. For brevity, the $S$ superscript is omitted in subsequent sections. While sensors often exhibit bias drift, we assume the system is stabilized after a warm-up period. So, the proposed algorithm does not consider biases in line with the author's previous work \cite{b9}. 

\subsection{Kalman Filter Process Model}
The system state is propagated using the following discrete-time transition:
\begin{equation}
\label{eq7}
\mathbf{x}_t^-=\boldsymbol{\Phi}_{t-1}\mathbf{x}_{t-1}+\mathbf{w}_{t-1},    
\end{equation}
$\boldsymbol{\Phi}$ is the state transition matrix, and $\mathbf{w}$ is the process model noise, which is white Gaussian with zero mean. This propagation's first-order approximation for the rotation matrix with gyroscope measurements is shown as,
\begin{equation}
\label{eq8}
\boldsymbol{\Omega}_t=\boldsymbol{\Omega}_{t-1}\left(\mathbf{I}_\mathbf{3}+\Delta t{\widetilde{\boldsymbol{\omega}}}_{t-1}\right).    
\end{equation}
Here, $\Delta t$ denotes the sampling interval, and ${\widetilde{\boldsymbol{\omega}}}_{t-1}$ represents the skew-symmetric matrix constructed from the gyroscope measurements at time $t-1$. Based on \eqref{eq8}, the a priori state vector propagates according to:
\begin{equation}
\label{eq9}
\mathbf{x}_t^-=\left(\mathbf{I}_\mathbf{3}-\Delta t{\widetilde{\mathbf{y}}}_{G,t-1}\right)\mathbf{x}_{t-1}+\Delta t(-{\widetilde{\mathbf{x}}}_{t-1})\mathbf{n}_G.    
\end{equation}
From this relationship, we can extract the state transition matrix, and the process noise vector, as follows:
\begin{equation}
\label{eq10}
\boldsymbol{\Phi}_{t-1}=\mathbf{I}_\mathbf{3}-\Delta t{\widetilde{\mathbf{y}}}_{G,t-1},    
\end{equation}
\begin{equation}
\label{eq11}
\mathbf{w}_{t-1}=\Delta t(-{\widetilde{\mathbf{x}}}_{t-1})\mathbf{n}_G.    
\end{equation}
Applying the expected value to the process noise times its transpose yields the process noise covariance matrix:
\begin{equation}
\label{eq12}
\mathbf{Q}_{t-1}=E[\mathbf{w}_{t-1}{\mathbf{w}^T}_{t-1}]=-\Delta t^2{\widetilde{\mathbf{x}}}_{t-1}\boldsymbol{\Sigma}_G{\widetilde{\mathbf{x}}}_{t-1}.
\end{equation}
In this expression, $\boldsymbol{\Sigma}_G$ stands for the gyroscope measurement noise covariance. Assuming isotropic noise characteristics with variance $\sigma_G^2$ across all three axes, this matrix is defined as $\boldsymbol{\Sigma}_G = \sigma_G^2\mathbf{I}_3$.

\subsection{Kalman Filter Measurement Model}
The system utilizes the standard measurement model:
\begin{equation}
\label{eq13}
\mathbf{z}_t=\mathbf{H}\mathbf{x}_t+\mathbf{v}_t,
\end{equation}
where $\mathbf{z}_t$ represents the measurement vector, $\mathbf{H}$ is the observation matrix, and $\mathbf{v}_t$ denotes the zero-mean white Gaussian measurement noise. To isolate the components of this model, we decouple the acceleration term from \eqref{eq7} into two distinct equations:
\begin{equation}
\label{eq14}
\mathbf{a}_t^-=\mathbf{a}_{t-1},    
\end{equation}
\begin{equation}
\label{eq15}
\boldsymbol{\varepsilon}_t=\mathbf{a}_t-\mathbf{a}_t^-.
\end{equation}
Replacing the body-frame gravity vector with the product of the states and norm of the gravity vector ($g = 9.81 \text{ m/s}^2$), we restructure \eqref{eq6} as:
\begin{equation}
\label{eq16}
\mathbf{y}_{\mathbf{A},{t}}-\mathbf{a}_{t-1}=g\mathbf{x}_t+\boldsymbol{\varepsilon}_t+\mathbf{n}_A.    
\end{equation}
Matching \eqref{eq16} to \eqref{eq13} establishes our model parameters:
\begin{equation}
\label{eq17}
\mathbf{z}_t=\mathbf{y}_{\mathbf{A},{t}}-\mathbf{a}_{t-1},
\end{equation}
\begin{equation}
\label{eq18}
\mathbf{H}=g\mathbf{I}_\mathbf{3},  
\end{equation}
\begin{equation}
\label{eq19}
\mathbf{v}_t=\boldsymbol{\varepsilon}_t+\mathbf{n}_A.
\end{equation}
The uncorrelated noise terms $\boldsymbol{\varepsilon}_t$ and $\mathbf{n}_A$ form the noise covariance matrix of the measurement:
\begin{equation}
\label{eq20}
\mathbf{M}_t=E[\mathbf{v}_t\mathbf{v}_t^T]=\boldsymbol{\Sigma}_{acc}+\mathbf{R}_A,
\end{equation}
Assuming isotropic accelerometer noise, $\mathbf{R}_A$ is modeled as $\sigma_A^2\mathbf{I}_3$. However, the time-varying modeling error covariance, $\boldsymbol{\Sigma}_{acc}$, cannot be calculated analytically because the true external acceleration is unknown during sampling. We address this issue in the following methodology.

\section{METHODOLOGY}
This section outlines the approach for tuning the measurement noise covariance matrix. The proposed algorithm employs an adaptive tuning strategy with innovation tracking, inspired by the concepts in \cite{b10,b11}, allowing the filter to dynamically adjust to unmodeled external accelerations. Within the Kalman filtering formulation, the innovation sequence, $\mathbf{e}_t$, is represented as 
\begin{equation}
\label{eq21}
\mathbf{e}_t=\mathbf{z}_t-\mathbf{H}\mathbf{x}_t^-,
\end{equation}
with $\mathbf{x}_t^-$ the prior state estimate, form the \textit{a priori} distribution after the prediction step. Whenever a measurement is heavily influenced by an outlier realization of the process noise, the predicted measurement covariance must be appropriately accounted for in the covariance evaluation to avoid a measurement update that is too confident in the observations, which would drive the estimate to high errors. Indeed, when external accelerations or disturbances occur, the magnitude of the innovation vector increases sharply, and the Kalman gain must be reduced accordingly. Therefore, the innovation covariance has an added term that increases its trace whenever needed: 
\begin{equation}
\label{eq22}
{\hat{\mathbf{C}}}_{e_t}=\mathbf{H}\mathbf{P}_t^-\mathbf{H}^T+{\hat{\boldsymbol{\Sigma}}}_{acc}+\mathbf{R}_A, 
\end{equation}
Thus, the denominator of the Kalman gain increases when poor measurements, such as outliers with high residual, are obtained:
\begin{equation}
\label{eq23}
\mathbf{K}_t=\mathbf{P}_t^-\mathbf{H}^T{(\mathbf{H}\mathbf{P}_t^-\mathbf{H}^T+{\hat{\boldsymbol{\Sigma}}}_{acc}+\mathbf{R}_A)}^{-1},    
\end{equation}
with $\mathbf{P}_t^-$ being the prior covariance matrix obtained after time propagation, while ${\hat{\boldsymbol{\Sigma}}}_{acc}$ indicates the best estimate for the $\boldsymbol{\Sigma}_{acc}$ matrix given in \eqref{eq20}. The true unknown measurement covariance ${\hat{\mathbf{C}}}_{e_t}$ must be estimated recursively at each measurement acquisition by inferring its value from the residuals. Indeed, after reaching steady-state operations, if the trace of the squared residual breaks the threshold
\begin{equation}
\label{eq24}
tr({{\mathbf{e}}}_t{{\mathbf{e}}}_t^T)\geq tr(\mathbf{H}\mathbf{P}_t^-\mathbf{H}^T+{\hat{\boldsymbol{\Sigma}}}_{acc}+\mathbf{R}_A).
\end{equation}
evaluated theoretically from the Kalman filtering formulation, it means there is an additional external acceleration acting on the measurement model, and the filtering process must be done in an adaptive fashion to counteract the higher noise level. To guarantee a rapid response to dynamic disturbances, the detection condition utilizes only the instantaneous measurement residual at time $t$, rather than a time-averaged innovation covariance.

\subsection{Multiple Tuning Factor Adaptation}
The innovation covariance can be redefined as
\begin{equation}
{\hat{\mathbf{C}}}_{e_t}=\frac{1}{\mu}\sum_{j=k-\mu+1}^{k}{\mathbf{e}_j\mathbf{e}_j^T},
\end{equation}
which indicates an approximation of the expected value operator via a finite sum of samples. This formulation can be restated to write the condition for estimating the ${\hat{\boldsymbol{\Sigma}}}_{acc}$,
\begin{equation}
\frac{1}{\mu}\sum_{j=k-\mu+1}^{k}{\mathbf{e}_j\mathbf{e}_j^T}=\mathbf{H}\mathbf{P}_t^-\mathbf{H}^T+\mathbf{S}_t+\mathbf{R}_A.    
\end{equation}
Defining $\mathbf{S}_t$ as a tuning matrix of MTFs and $\mu$ as the innovation covariance moving window size (fixed at 1 herein), we can rearrange the expression to solve for $\mathbf{S}_t$:
\begin{equation}
\mathbf{S}_t=\ \frac{1}{\mu}\sum_{j=k-\mu+1}^{k}{\mathbf{e}_j\mathbf{e}_j^T}-\mathbf{H}_t\mathbf{P}_t^-\mathbf{H}_t^T-\mathbf{R}_A.    
\end{equation}
When the condition in \eqref{eq24} holds, the elements of $\mathbf{S}_t$ corresponding to the accelerated axes dynamically adjust, thereby efficiently tuning the measurement noise covariance. The final acceleration covariance estimate, ${\hat{\boldsymbol{\Sigma}}}_{acc}$, is thus given by:
\begin{equation}
{\hat{\boldsymbol{\Sigma}}}_{acc}=diag(s_1,s_2,s_3),    
\end{equation}
\begin{equation}
s_i=max\left\{0,{\mathbf{S}}_{ii}\right\}, i=1,2,3. 
\end{equation}
Here, ${\mathbf{S}}_{ii}$ is the $i$-th diagonal entry of $\mathbf{S}$. During external disturbances, $\mathbf{S}$ inflates the covariance matrix appropriately; absent such acceleration, the diagonal terms revert to zero for standard operation.

\subsection{Jerk-Augmented EKF Architecture}
To address the high-frequency disturbances inherent in agricultural environments, we implement an EKF with a jerk-augmented state space. This formulation allows the filter to account for the rate of change of acceleration, providing robustness against sudden terrain-induced shocks and uneven soil densities.

\subsubsection{State Vector and Input}
The state vector $\mathbf{x}_k \in \mathbb{R}^{9}$ comprises the 3D position $\mathbf{p}$, velocity $\mathbf{v}$, and jerk $\mathbf{j}$ in the East-North-Up (ENU) frame:
\begin{equation}
\mathbf{x}_k = \begin{bmatrix} \mathbf{p}_k^T & \mathbf{v}_k^T & \mathbf{j}_k^T \end{bmatrix}^T
\end{equation}
The control input $\mathbf{u}_k$ is the specific force measured by the IMU, transformed into the ENU frame and gravity-compensated:
\begin{equation}
\mathbf{u}_k = \boldsymbol{\Omega}\mathbf{a}_{body} + \mathbf{g}_{enu}
\end{equation}

\subsubsection{Time Update (Prediction)}
The state is propagated using a kinematic model where the input acceleration drives position and velocity, while jerk is modeled as a random walk process. The discrete-time prediction equations are:
\begin{equation}
\mathbf{x}_{k|k-1} = \mathbf{F}_k \mathbf{x}_{k-1|k-1} + \mathbf{B}_k \mathbf{u}_k
\end{equation}
\begin{equation}
\mathbf{P}_{k|k-1} = \mathbf{F}_k \mathbf{P}_{k-1|k-1} \mathbf{F}_k^T + \mathbf{Q}_k
\end{equation}
Matching the third-order Taylor expansion for position, the state transition matrix $\mathbf{F}_k$ and input matrix $\mathbf{B}_k$ are defined as:
\begin{equation}
\mathbf{F}_k = \begin{bmatrix} 
\mathbf{I}_3 & \Delta t \mathbf{I}_3 & \frac{\Delta t^3}{6} \mathbf{I}_3 \\
\mathbf{0}_3 & \mathbf{I}_3 & \frac{\Delta t^2}{2} \mathbf{I}_3 \\
\mathbf{0}_3 & \mathbf{0}_3 & \mathbf{I}_3 
\end{bmatrix}, \quad
\mathbf{B}_k = \begin{bmatrix} 
\frac{\Delta t^2}{2} \mathbf{I}_3 \\
\Delta t \mathbf{I}_3 \\
\mathbf{0}_3 
\end{bmatrix}
\end{equation}

\subsubsection{Measurement Update (Correction)}
The filter updates the state estimate using a GNSS position and odometry velocity measurements. The measurement vector is $\mathbf{z}_k = [\mathbf{p}_{gnss}^T, \mathbf{v}_{gnss}^T]^T$. The update equations are:
\begin{align}
\mathbf{y}_k &= \mathbf{z}_k - \mathbf{H} \mathbf{x}_{k|k-1} \\
\mathbf{S}_k &= \mathbf{H} \mathbf{P}_{k|k-1} \mathbf{H}^T + \mathbf{R}_{gnss, k} \\
\mathbf{K}_k &= \mathbf{P}_{k|k-1} \mathbf{H}^T \mathbf{S}_k^{-1} \\
\mathbf{x}_{k|k} &= \mathbf{x}_{k|k-1} + \mathbf{K}_k \mathbf{y}_k \\
\mathbf{P}_{k|k} &= \mathbf{P}_{k|k-1} - \mathbf{K}_k \mathbf{S}_k \mathbf{K}^T_k
\end{align}
The observation matrix $\mathbf{H}$ maps the state to the available position and velocity measurements, leaving jerk unobserved:
\begin{equation}
\mathbf{H} = \begin{bmatrix} 
\mathbf{I}_3 & \mathbf{0}_3 & \mathbf{0}_3 \\
\mathbf{0}_3 & \mathbf{I}_3 & \mathbf{0}_3 
\end{bmatrix}
\end{equation}
Critically, the measurement noise covariance $\mathbf{R}_{gps, k}$ is constructed as a block-diagonal matrix:
\begin{equation}
\mathbf{R}_{gnss, k} = \text{blkdiag}(\mathbf{R}_{pos, k}, \mathbf{R}_{vel, k})
\end{equation}
Here, $\mathbf{R}_{pos, k}$ and $\mathbf{R}_{vel, k}$ are the time-varying covariances provided by the GNSS receiver indicating signal quality.

\section{EXPERIMENTAL SETUP}
To validate the efficacy of the proposed navigation framework, a series of field experiments were conducted using Salin247's commercial autonomous agricultural platform. This section details the hardware configuration, the data acquisition process in unstructured farm environments, and the quantitative metrics used to benchmark the proposed jerk-augmented MTF-EKF against the standard baseline method.

\subsection{Experimental Hardware} Field validation was conducted using an autonomous agricultural robot developed by Salin247. The platform is a four-wheel independent drive (4WID) electric vehicle designed for precision tasks such as seeding, spraying, planting, and crop monitoring in unstructured off-road farm environments. The primary navigation sensor suite consists of a Fixposition Vision-RTK 2 sensor, which provides:
\begin{itemize}
\item Global positioning via dual-antenna GNSS.
\item Inertial measurements from an internal IMU.
\item Visual odometry cues (fused internally by the sensor).
\end{itemize}
Ground truth for validation was established using the sensor's high-precision RTK-fixed solution fused with visual odometry, which offers centimeter-level accuracy under open-sky conditions.

\subsection{Data Collection and Processing}
Data was recorded using the Robot Operating System (ROS) middleware. The dataset, stored in \texttt{.bag} format, captures the vehicle executing typical agricultural maneuvers, including straight-line tracking on uneven soil and sharp headland turns. Key topics extracted for post-processing include:
\begin{itemize}
\item \texttt{/fixposition/poiimu}: Raw angular velocity and linear acceleration sampled at 100 Hz.
\item \texttt{/fixposition/gnss1}: Latitude, longitude, and covariance estimates for the measurement update steps.
\item \texttt{/fixposition/odometry}: Used as the reference ground truth for error analysis.
\end{itemize}
The proposed jerk-augmented MTF-EKF was implemented in MATLAB and compared against a baseline 6-state (Position-Velocity) EKF. Both filters utilized the same IMU inputs and GNSS measurement updates to ensure a fair comparison of the kinematic modeling effects.

\subsection{Performance Metrics}
To evaluate the resilience of the proposed method, we computed the RMSE for both the 3-D position and 3-D velocity estimates over a 150-second test trajectory. The total 3-D RMSE is defined as:
\begin{equation}RMSE_{pos} = \sqrt{\frac{1}{N} \sum_{k=1}^{N} |\mathbf{p}_{est,k} - \mathbf{p}_{true,k}|^2}\end{equation}

\section{RESULTS}
\begin{figure}[htbp]
\centerline{\includegraphics[width=0.5\textwidth]{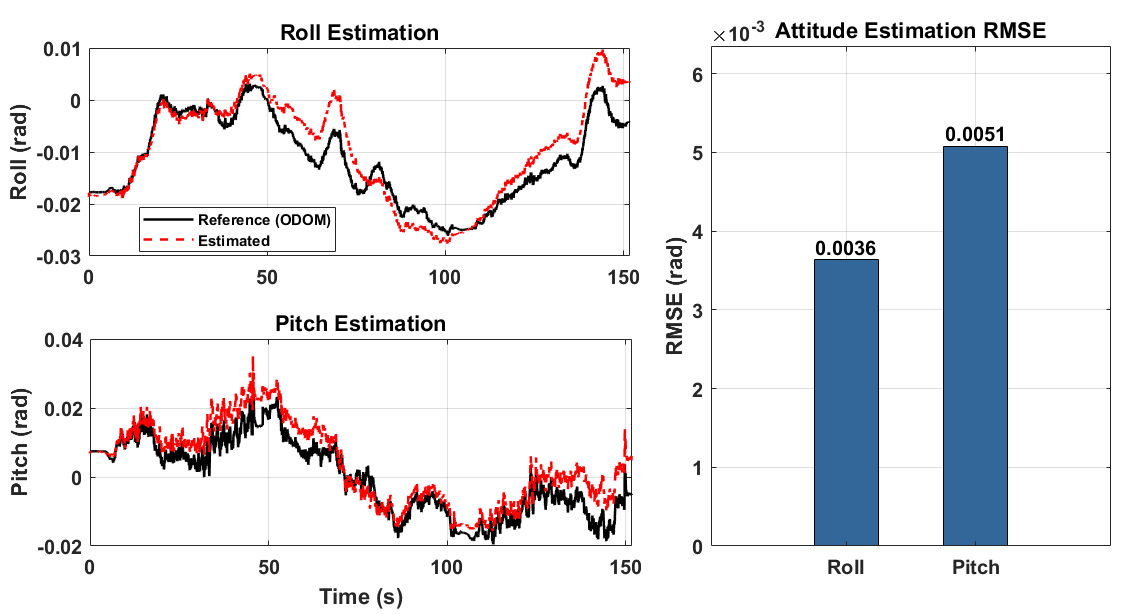}}
\caption{Estimation performance of tilt angles using MTF methodology}
\label{fig4}
\end{figure}

\begin{figure}[htbp]
\centerline{\includegraphics[width=0.5\textwidth]{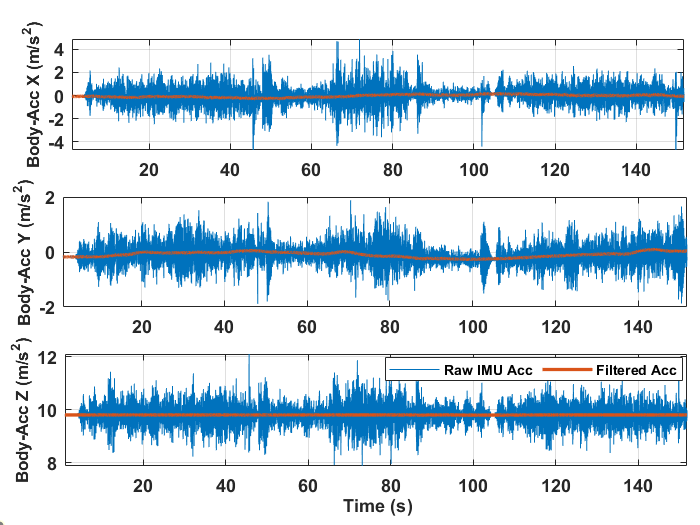}}
\caption{Visual representation of external acceleration compensation}
\label{fig6}
\end{figure}

The performance of the MTF methodology in the adaptive Kalman filter is illustrated in Fig. \ref{fig4}. The left column displays the time history of the estimated roll and pitch angles compared against the ground truth solution. The filter demonstrates stable tracking of the vehicle's tilt, even during the high-vibration intervals characteristic of agricultural terrain. The quantitative accuracy is summarized in the bar chart (Fig. \ref{fig4}, right), showing RMSE of {0.0051 radians} for roll and {0.0076 radians} for pitch. These low error margins confirm that the gravity compensation mechanism effectively isolates the specific force vector, providing a reliable orientation estimate for the subsequent navigation stage. This capability is further visualized in Fig. \ref{fig6}, which compares the raw accelerometer measurements against the estimated body acceleration, which is compensated against external acceleration/disturbances. As shown, the adaptive filter successfully decouples the high-frequency dynamic motion from the quasi-static gravity vector.

\begin{figure}[htbp]
\centerline{\includegraphics[width=0.5\textwidth]{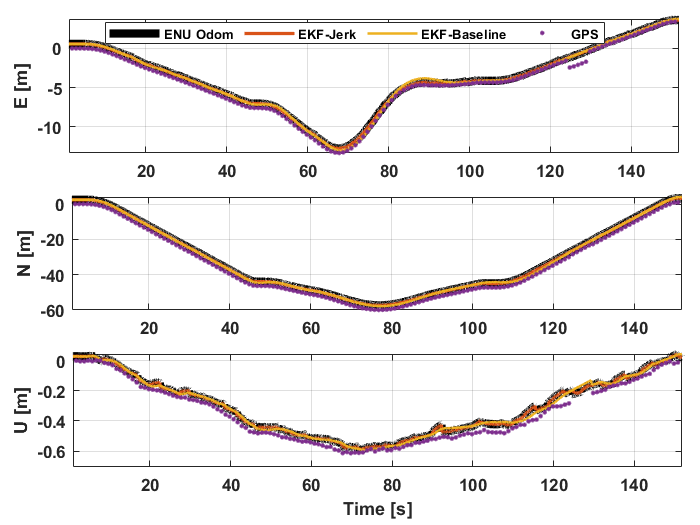}}
\caption{Position error norm analysis}
\label{fig1}
\end{figure}

\begin{figure}[htbp]
\centerline{\includegraphics[width=0.5\textwidth]{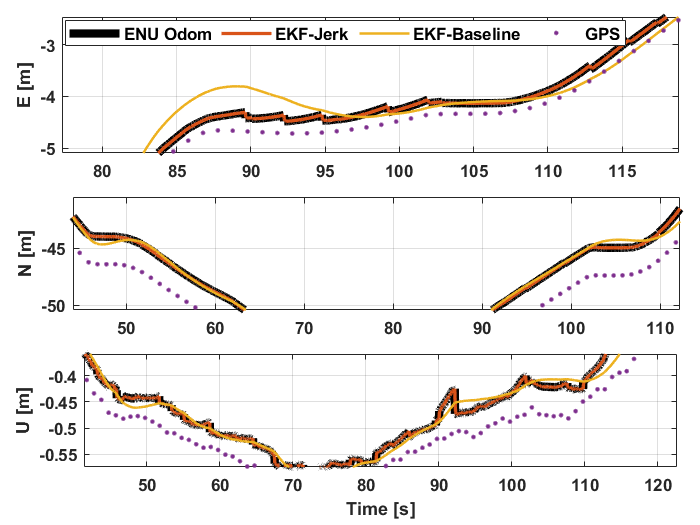}}
\caption{Close-up view of position error norm analysis}
\label{fig2}
\end{figure}

\begin{figure}[htbp]
\centerline{\includegraphics[width=0.5\textwidth]{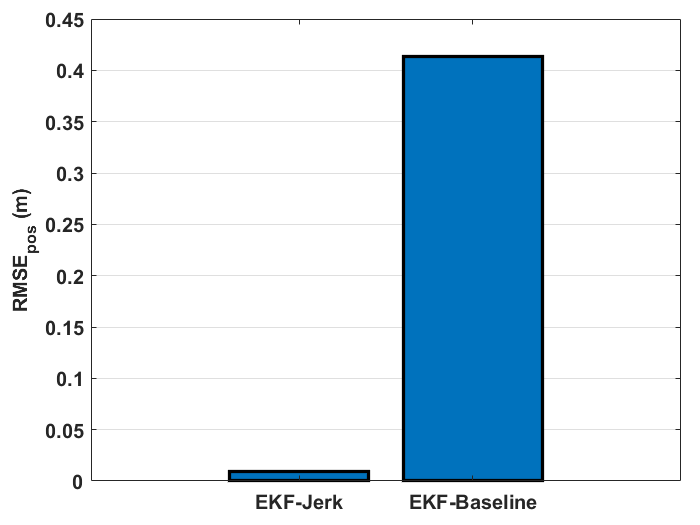}}
\caption{Bar chart RMSE summary for EKF methodologies}
\label{fig3}
\end{figure}

The performance of the proposed method can be attributed to the jerk state's ability to absorb high-frequency process noise. As shown in the position error norm analysis (Fig. \ref{fig1}), the baseline EKF exhibits ``spikes" in position error whenever the robot encounters a surface irregularity (e.g., at $t=45$s and $t=90$s as shown in Fig. \ref{fig2}). In contrast, the jerk-augmented filter effectively predicts the rate of change of acceleration, allowing the MTF adaptation to distinguish between sensor noise and true vehicle dynamics. This results in a smoother estimated trajectory that adheres closer to the RTK ground truth, confirming the hypothesis that higher-order kinematic modeling is essential for robust agricultural navigation. Finally, Fig. \ref{fig3} summarizes the importance of jerk augmentation with MTF adaptation, showing that EKF with jerk augmentation achieves
centimeter-level positional accuracy.

\begin{figure}[htbp]
\centerline{\includegraphics[width=0.5\textwidth]{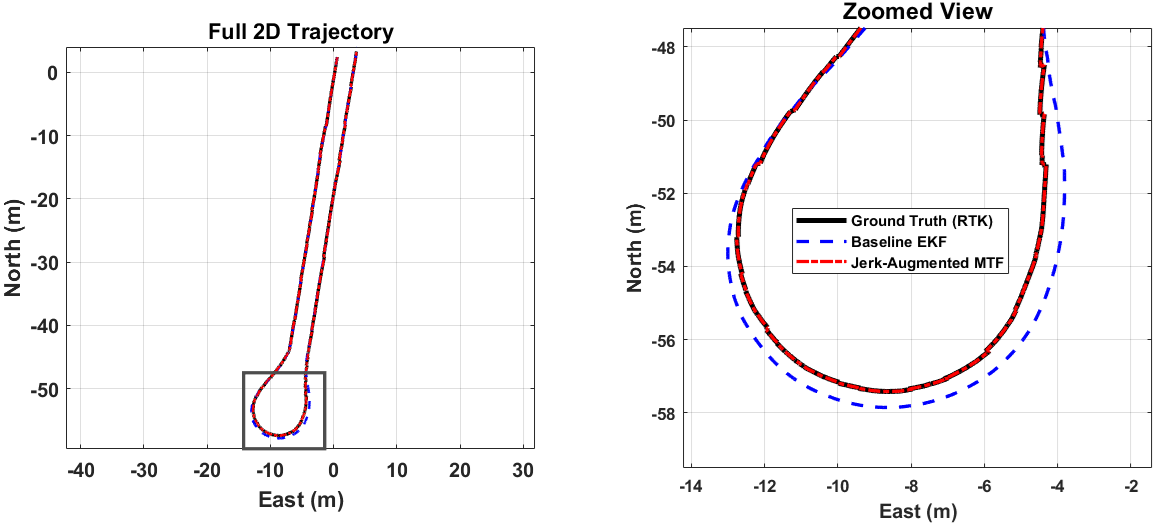}}
\caption{ 2-D trajectory estimation performance comparison between baseline and jerk-augmented EKFs}
\label{fig5}
\end{figure}

Fig. \ref{fig5} presents the 2-D trajectory estimation results in the ENU frame. The left panel shows the global path of the Salin247 robot during a 150-second field test. While the baseline EKF (blue dashed line) drifts noticeably during turns and uneven terrain segments, the jerk-augmented MTF-EKF (red dash-dot line) maintains a tight lock on the ground truth trajectory (black solid line). The zoomed inset in Fig. \ref{fig5} highlights a specific instance of disturbance rejection. At these coordinates, the robot encountered a surface irregularity that caused a sudden acceleration spike. The baseline filter, constrained by its constant-acceleration assumption, overshot the turn. In contrast, the jerk-augmented filter successfully absorbed the impulsive force, resulting in a trajectory that is physically consistent with the robot's actual motion. This superior tracking capability is further validated by the reduction in total position RMSE shown in the accompanying bar chart.

\section{CONCLUSION}
This paper presented a robust navigation framework for autonomous agricultural robots operating in harsh environments that might face sensor outages. By augmenting the standard kinematic state space with a jerk process model, the proposed filter effectively captures the high-frequency dynamics and impulsive forces inherent in off-road terrain. Furthermore, the integration of a novel MTF adaptive mechanism allows the system to adjust the measurement noise covariance, selectively isolating sensor axes corrupted by vibration or impact.

Experimental validation on the Salin247 platform demonstrated that the jerk-augmented MTF-EKF reduces 3-D position RMSE more than ten times compared to a baseline EKF. The results confirm that accounting for the rate of change of acceleration is critical for minimizing lag and drift during dynamic maneuvers. Future work will focus on integrating visual odometry cues into the MTF framework to further enhance resilience during extended sensor outages. Moreover, coupling these robust state estimates with advanced control strategies, such as model predictive controllers, enables high-precision trajectory tracking and active disturbance rejection in closed-loop operation.

\section*{ACKNOWLEDGEMENT}
The authors would like to thank Salin247 Inc. for supporting this research. We specifically extend our gratitude to the engineering team for providing access to the autonomous agricultural robot platform and for their assistance with the field data collection. This collaboration was instrumental in
validating the proposed algorithms in a real-world agricultural
environment.

The work of this project has been funded by the Seed Grant Award for Digital and Precision Agriculture at Iowa State University under grant number PG114523.

\addtolength{\textheight}{-12cm}   




\end{document}